\documentclass[10pt,twocolumn,letterpaper]{article}
\usepackage{cvpr}
\usepackage{times}
\usepackage{helvet}
\usepackage{courier}

\usepackage{pgfplots}
\usepackage{amsmath}
\usepackage{rotating}
\usepackage{graphicx}
\usepackage{multirow}
\usepackage{amsthm}
\usepackage{amsfonts}
\usepackage{amssymb}
\usepackage{epsfig}
\usepackage{bbm}

\usepackage{tikz}

\usepackage{pifont}

\usetikzlibrary{shapes.geometric}
\usetikzlibrary{backgrounds}
\usetikzlibrary{fit}
\usepackage[breaklinks=true,bookmarks=false]{hyperref}

\cvprfinalcopy 

\def\cvprPaperID{****} 

\begin{document}

\title{Weakly Supervised Tracklet Person Re-Identification by Deep Feature-wise Mutual Learning}

\author{Zhirui Chen\textsuperscript{\rm{1,2}}, Jianheng Li\textsuperscript{\rm{1,3}}, Wei-Shi Zheng\textsuperscript{\rm {1,3}}\thanks{Corresponding author.} \\
\textsuperscript{\rm 1}School of Data and Computer Science, Sun Yat-sen University, China\\ 
\textsuperscript{\rm 2}Accuvision Technology Co.,LTD\\
\textsuperscript{\rm 3}Key Laboratory of Machine Intelligence and Advanced Computing, Ministry of Education, China\\
chenzhr5@mail2.sysu.edu.cn, lijheng3@mail2.sysu.edu.cn, wszheng@ieee.org
}


\maketitle

\begin{abstract}
\begin{quote}
The scalability problem caused by the difficulty in annotating Person Re-identification(Re-ID) datasets has become a crucial bottleneck in the development of Re-ID.
To address this problem, many unsupervised Re-ID methods have recently been proposed.
Nevertheless, most of these models require transfer from another auxiliary fully supervised dataset, which is still expensive to obtain.
In this work, we propose a Re-ID model based on Weakly Supervised Tracklets(WST) data from various camera views, which can be inexpensively acquired by combining the fragmented tracklets of the same person in the same camera view over a period of time.
We formulate our weakly supervised tracklets Re-ID model by a novel method, named deep feature-wise mutual learning(DFML), which consists of Mutual Learning on Feature Extractors (MLFE) and Mutual Learning on Feature Classifiers (MLFC).
We propose MLFE by leveraging two feature extractors to learn from each other to extract more robust and discriminative features.
On the other hand, we propose MLFC by adapting discriminative features from various camera views to each classifier. Extensive experiments demonstrate the superiority of our proposed DFML over the state-of-the-art unsupervised models and even some supervised models on three Re-ID benchmark datasets.
\end{quote}
\end{abstract}

\section{Introduction}
\label{sec:introduction}

The objective of person re-identification(Re-ID) is to recognize people across non-overlapping surveillance camera views. Benefiting from deep learning in recent years, the performance of Re-ID models has achieved significant progress~\cite{ahmed2015improved,SVDNet,PCB}.
However, a crucial bottleneck of current person Re-ID is the limited scalability problem due to the fully supervised annotation Re-ID datasets. In the context of conventional supervised Re-ID, we need to annotate person bounding box images and label the positive or negative image pairs across different camera views. It is quite expensive for the reason that (1) we have no knowledge of where and when the person that occurred in one camera view occurs in other camera views, and (2) human annotators are hard to remember such a large number of person images. 

\begin{figure}[!t]
    \centering
    \includegraphics[width=0.45\textwidth]{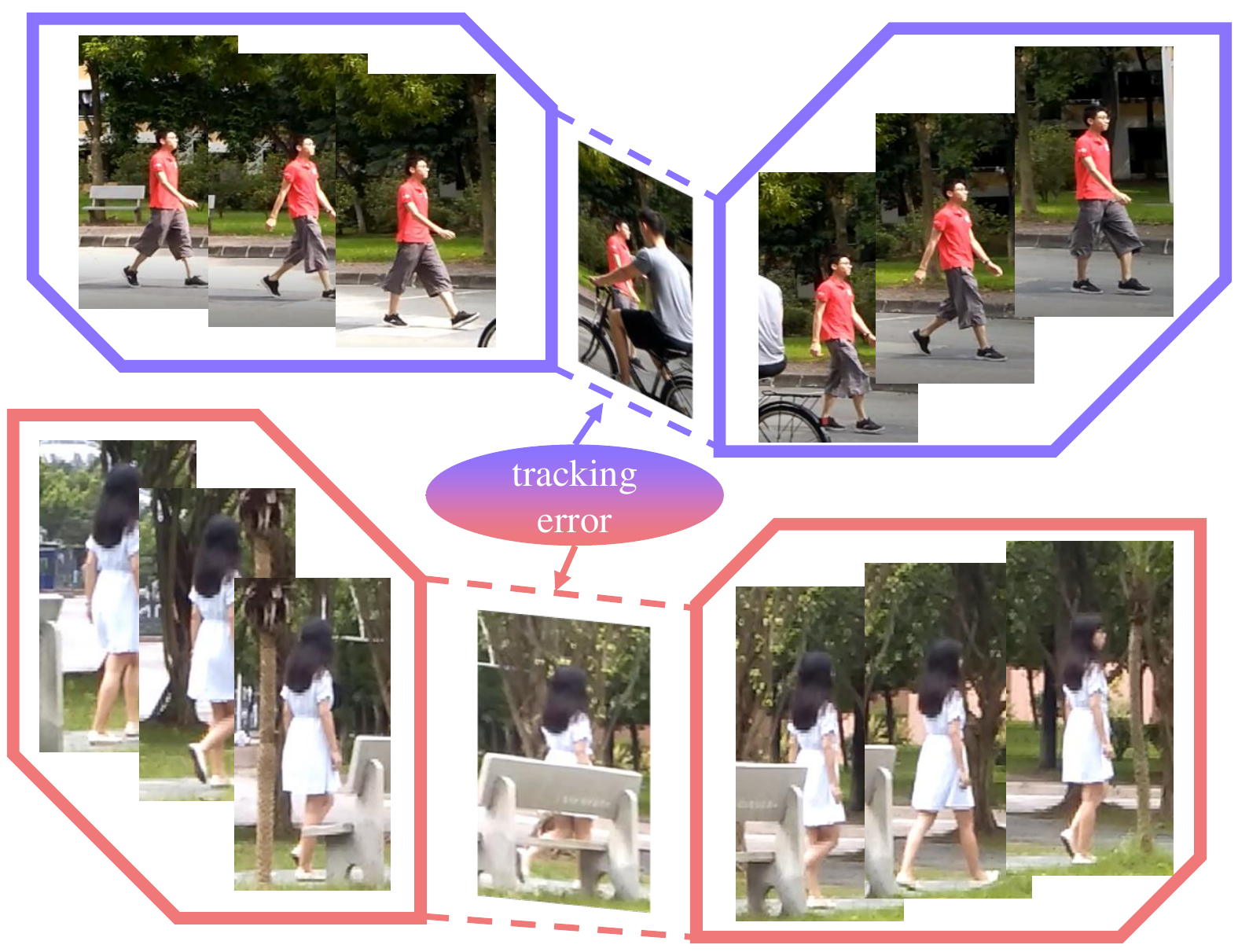}
    \caption{Tracklets separate because of tracking error, which can be inexpensively combined to WST by human annotators.}
    \label{fig:tracklet}
\end{figure}

To address the scalability problem in Re-ID, many recent works have focused on unsupervised person re-identification~\cite{PUL,CAMEL,Gong_unsupervised}. However, existing unsupervised learning Re-ID methods are not quite competitive compared with the supervised Re-ID methods, because it is difficult to learn robust and discriminative features in the absence of pairwise label guidance. 

In this paper, we consider a weakly supervised person Re-ID learning problem, that leverages the weakly supervised person tracklets data from a surveillance video. As is shown in Fig.~\ref{fig:tracklet}, it is inevitable that raw person tracklets would be fragmented even if they belong to the same person because of occlusion, motion blur, scale variation, and other tracking errors. Thus, we refer to the weakly supervised tracklets (WST) by combining the raw person tracklets, which belong to the same person during annotation. This annotation is quite inexpensive, because in most cases we only combine the raw person tracklets within a very short time on the same camera view. 

Furthermore, we leverage WST to our deep feature-wise mutual learning. First, we propose a ``two feature extractors'' architecture design where two networks learn from each other mutually. Second, we propose a multiple feature classifiers architecture design where learning information delivers among classifiers so that they can learn from each other.

We summarize our contributions as follows:
\begin{enumerate}
    \item We refer to weakly supervised tracklets (WST) by combining the raw tracklets from the same person, within a narrow space-time area.
    \item We formulate our deep feature-wise mutual learning (DFML) by way of Mutual Learning on Feature Extractors (MLFE) and Mutual Learning on Feature Classifiers (MLFC).
    \item Our proposed DFML achieves competitive performance in three widely used benchmarks, and ablation experiments demonstrate the effectiveness of MLFE and MLFC. 
\end{enumerate}

\section{Related Work}
\label{sec:releated_work}

\subsection{Person Re-ID}
\label{sec:rw_reid}

\subsubsection{Supervised Re-ID Learning}
\label{sec:rw_sup_reid}
With the development of deep learning in recent years, significant improvements have been made in the person Re-ID field, specifically, supervised person Re-ID.
In the early years, because of the limitation of the size of person Re-ID datasets, researchers tended to train a pairwise siamese network~\cite{siamese1,siamese2}, where a binary-classification model is optimized with a pair of person images as input and a probability is output that indicated whether the two images belong to the same person.
Later, with the increasing scale of datasets, many works~\cite{MARS,market1501,PRW} treated the training of person Re-ID as an image classification problem, in which all images of the same person are regarded as the same class.

\subsubsection{Unsupervised Re-ID Learning}
\label{sec:rw_unsup_reid}
Recently, supervised person Re-ID has encountered several tricky issues, such as expensive annotations of persons and domain adaptation problems.
These problems led to the appearance of a special type of unsupervised person Re-ID, where a model is trained on a labeled dataset(source domain), and then knowledge is transferred to another unlabeled dataset(target domain).
Some methods~\cite{PUL,CAMEL} adopted clustering to assign a pseudo label to unlabeled person images in the self-training framework, while others~\cite{SPGAN,PTGAN} used GAN to generate target images with a similar style of images from the source domain.
Li et al.~\cite{Gong_unsupervised} began with the special temporal and spatial characteristic of person Re-ID, assuming that the positive tracklet pairs are not available, even on the same camera view. Our motivation is similar to theirs, but we assume that for the same person under the same camera view, a tracklet may be truncated into multiple tracklets due to tracking errors. These errors are easily corrected by a human annotator.

\subsubsection{Weakly Supervised Re-ID Learning}
\label{sec:rw_weak_reid}
Several works focused on the weakly supervised learning for person Re-ID.
In person search, a sub field of person Re-ID, the gallery is a raw scene image, rather than the cropped bounding box. To some extent, it is a weakly supervised setting. However, methods of person search require full supervised annotation for the training of person detection. The setting proposed by Meng et al.~\cite{weak_supervised1} is closer to weakly supervised learning, where we only access the multiple video-based person label, namely the persons occurred in the video, but their respective tracklets. In a similar way, Wang et al.~\cite{weak_supervised2} proposed a large weakly supervised Re-ID dataset where they replaced image-level annotations with bag-level annotations.

\subsection{Deep Mutual Learning}
\label{sec:rw_mutual}
Hinton et al. first introduced knowledge distillation~\cite{distillation}, transferring the knowledge from teacher networks(big networks) to student networks(small networks), to improve the performance of the student network.
However, Zhang et al. found that it doesn't require a small network as the student network. It also worked well on a student network with the same scale as the teacher network.
Therefore they proposed deep mutual learning, where two networks learn from each other at the same time~\cite{mutual}.
The method proposed in this work is similar to deep mutual learning. The main difference is that, deep mutual learning proposed by Zhang et al. is ``predict-wise'' learning while ours is ``feature-wise'' learning, which will be explained in details in the following section.

\begin{figure*}
    \centering
    \includegraphics[height=0.22\textheight, width=0.8\textwidth]{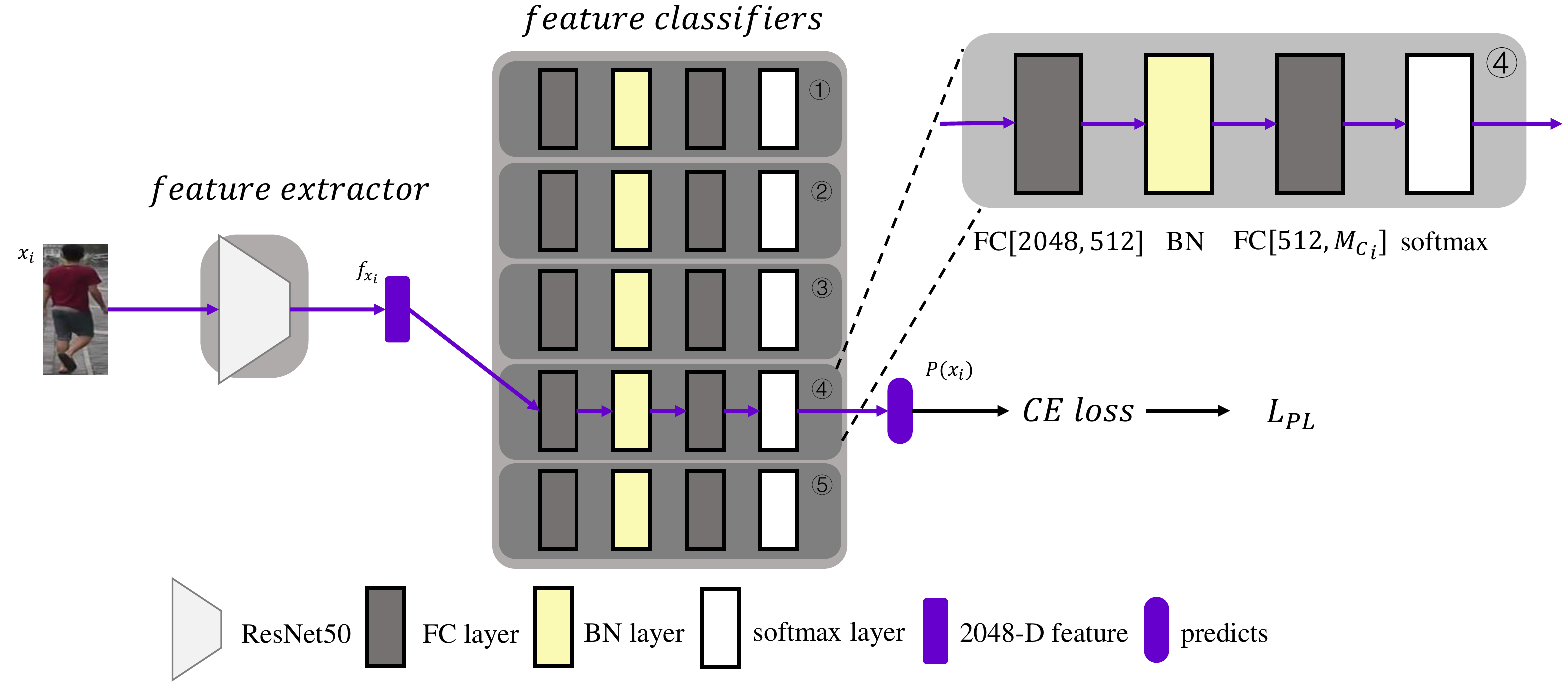}
    \caption{An overview of our Weakly Supervised Tracklet Re-ID baseline model. This is a case for 5 camera views which correspond to 5 classifier branches, respectively. A person image from the 4th camera view is fed into the feature extractor for obtaining the feature. Then, the feature is fed into the 4th branch for obtaining the predicts, which is used to compute $L_{PL}$.}
    \label{fig:model1}
\end{figure*}

\section{Proposed Method}
\label{sec:method}

\subsection{Problem Setting and Overview}
\label{sec:overview}

Let $\{x_i,y_i, c_i\}_{i=1}^{N}$ denotes the training set, where $x_i$ is a person bounding box image in WST labeled $y_i$ collected in camera view labeled $c_i$; $y_i = 1,\cdots,M_{c_i}$, where $M_{c_i}$ is the number of WST from camera view $c_i$; $c_i = 1,\cdots,C$, where $C$ is the number of camera views. In our weakly supervised tracklet formulation, raw person tracklets extracted from the same camera view, are combined into one WST during label annotation if they correspond to the same person. Thus, $y_i$ indicates an independent person label under a specific camera view $c_i$; therefore, we also call $y_i$ a person label in the remainder of this paper. In contrast to fully supervised Re-ID, in our weakly setting we are unaware of the mapping relation of the same person across various camera views in the training set, which is a crucial challenge in our problem setting. 

Our goal is to learn a deep discriminative feature representation model for Re-ID matching, which is the essential objective of Re-ID. In the following, to introduce our Deep Feature-wise Mutual Learning (DFML), we first introduce a baseline method.

\subsection{Our Baseline Model}
\label{sec:baseline}

Li et al. proposed Per-Camera Tracklet Discrimination (PCTD) learning~\cite{Gong_unsupervised}. Specifically, the model of PCTD formulates $C$ distinct classification tasks within $C$ camera views in a multi-branch network architecture design. These $C$ classification tasks share a ResNet-50 network backbone~\cite{resnet} as a feature extractor but have their individual fully connected layer as a classifier branch.

Inspired by PCTD, we propose our baseline model for Weakly Supervised Tracklet Person Re-Identification, which is more appropriate and more effective for our problem formulation.  

\subsubsection{Per-Camera Learning}
\label{sec:PCL}

In our baseline model, we formulate $C$ distinct classification tasks, which share a ResNet-50 network backbone and have their individual classifier branches. As is shown in Fig.~\ref{fig:model1}, we adopt two FC layers as a classifier branch, with a BN layer\cite{BN} followed by the first FC layer in each classifier branch. Then, the softmax Cross-Entropy (CE) loss function is used to optimize each classification task. The CE loss of the $i$-th sample is computed as follows:

\begin{equation}
    L_{ce}(x_i)=-logP_{y_i}(x_i)
\end{equation}
where $P_{y_i}(x_i) \in \mathbb{R}$ is the $y_i$-th element of $P(x_i)$, and $P(x_i) \in \mathbb{R}^{M_{c_i}}$ is the output of the softmax layer in classifier branch $c_i$, as shown in Fig.~\ref{fig:model1}.



Given a mini-batch $\chi$, we formulate the Per-Camera Learning as follows:
\begin{equation}
    L_{PL}^{0}= \frac{1}{\Vert \chi \Vert}\sum_{x_i \in \chi} L_{ce}(x_i)
\end{equation}

\subsubsection{Cross-Camera Learning}
\label{sec:CCL}

In the context of Re-ID, a person should only be discriminated by his visual appearance irrespective of the circumstances that he is in and the cameras that he is caught. Thus, the discriminative feature of the person image should be independent of the camera view. We assume that the distribution of people from different camera views is equivalent. It is difficult and evenly impossible to learn the features that meet the equivalent distribution of various camera view. Here we simply focus our attention on the restriction of features' arithmetic means. To this end, the arithmetic means of discriminative features obtained from each camera view should be pulled closer during the training phase. Thus, we formulate the Cross-Camera Learning as follows:        
\begin{equation}
	L_{CL}^{0}=\sum_{u\ne v} \frac{\Vert \mu_u - \mu_v \Vert_2^2}{C_{batch} \times (C_{batch}-1)}
\end{equation}
\begin{equation}
\label{equation:mu}
    \mu_u = \frac{1}{\Vert \chi_u \Vert}\sum_{x_i \in \chi_u} f_{x_i}
\end{equation}
where $f_{x_i} \in \mathbb{R}^{2048}$ is the discriminative feature extracted by the feature extractor; $\chi_u$ is the images set from camera view $u$ in a mini-batch; $\mu_u \in \mathbb{R}^{2048}$ is the mean of those features from camera view $u$ in a mini-batch; $C_{batch}$ is the number of camera views in a mini-batch, and $C_{batch} \times (C_{batch}-1)$ is a normalization factor. 


Based on the above analysis, we formulate our baseline model as follows:
\begin{equation}
	L_{baseline} = L_{PL}^{0} + \lambda L_{CL}^{0}
\end{equation}
where $\lambda$ is the weight coefficient to balance Per-Camera Learning and Cross-Camera Learning.

\begin{figure*}
    \centering
    \includegraphics[height=0.25\textheight, width=0.8\textwidth]{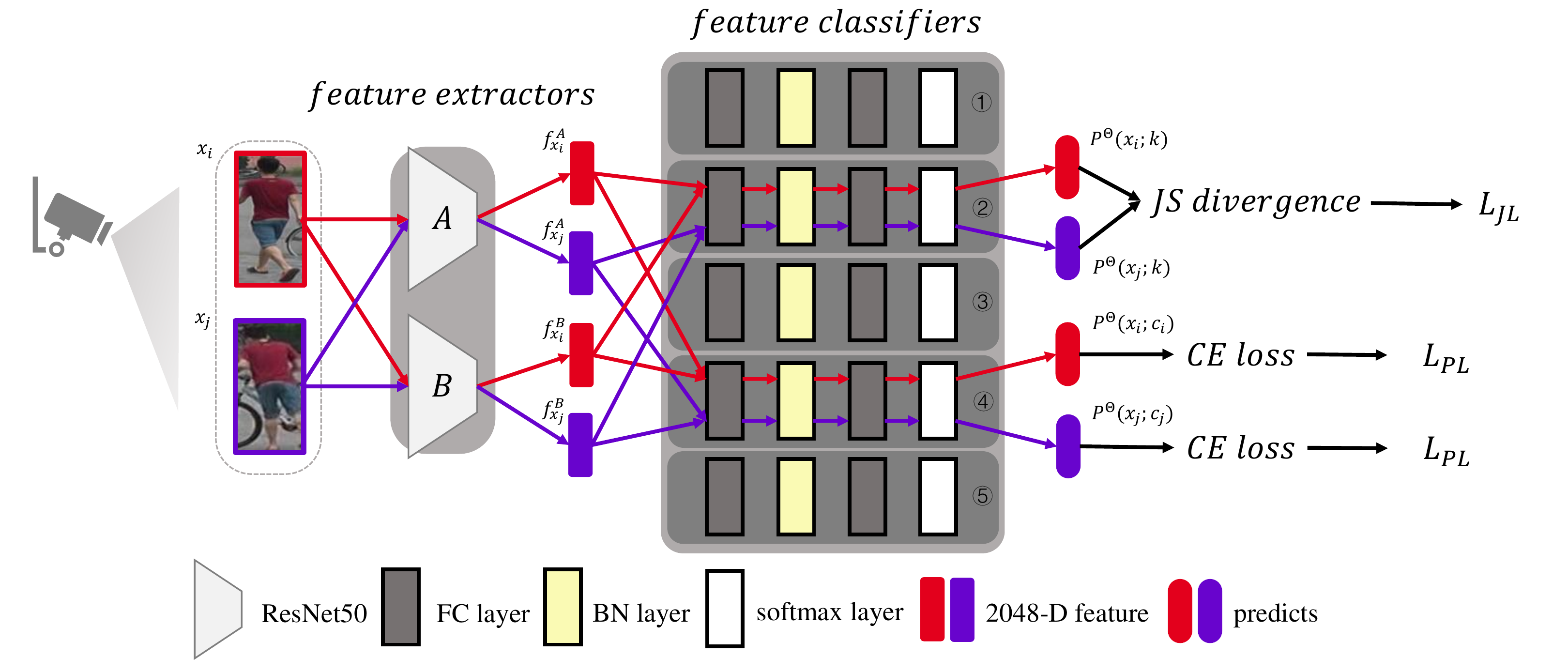}
    \caption{An overview of our DFML model. There are two images of the same person from the 4th camera view. The red stream corresponds to the forward propagation of the first image, while the purple stream corresponds to the second image. All of the extracted features are fed into the 4th branch to compute $L_{PL}$. Additionally, the features of the two images are fed into other branches(\textit{e.g.} the 2nd branch) to compute $L_{JL}$.}
    \label{fig:model2}
\end{figure*}

\subsection{Mutual Learning on Feature Extractors (MLFE)}
\label{sec:FEML}

In conventional deep mutual learning~\cite{mutual}, there are two individual networks, and these networks do not share any weight parameters. Zhang et al. proposed a loss function to compute the Jensen-Shannon Divergence of the two networks' predicts(generated after the softmax layer), where both of predicts are obtained from the same input image. During the back-propagation in training, the two networks' parameters would update depending on both networks' predicts. Consequently, the two networks learn from each other mutually, which achieves the deep mutual learning design. In Zhang's experiment, this design performed better than a single network design. In this way, deep mutual learning is guided by the predicts, which we call it ``predict-wise'' mutual learning.  

In contrast, our work achieves the deep mutual learning design in a ``feature-wise'' manner, which is guided by discriminative features. As shown in Fig.~\ref{fig:model2}, we also adopt the two networks design. Based on our aforementioned baseline model in Per-Camera Learning, we leverage two networks to extract discriminative features, and both networks are followed by the same cluster of $M_{c_i}$ classifier branches that correspond to the $M_{c_i}$ classification tasks. 
With two networks design, the predicts $P(x_i)$ is rewritten by $P^{\Theta}(x_i)$, as are $L_{ce}^{\Theta}$ and $L_{PL}^{\Theta}$. Specifically, they are formulated as follows:
\begin{equation}
    L_{ce}^{\Theta}(x_i)=-logP_{y_i}^{\Theta}(x_i)
\end{equation}

\begin{equation}
    L_{PL}^{\Theta}= \frac{1}{\Vert \chi \Vert}\sum_{x_i \in \chi} L_{ce}^{\Theta}(x_i)
\end{equation}
where $\Theta \in \{A,B\}$ is a network label, and $P_{y_i}^{\Theta}(x_i) \in \mathbb{R}$ is the $y_i$-th element of $P^{\Theta}(x_i)$.



In summary, We formulate our Mutual Learning on Feature Extractors as follows:
\begin{equation}
	L_{PL} = \frac{1}{2}(L_{PL}^{A} + L_{PL}^{B})
\end{equation}

Most conventional deep learning models for person Re-ID also consist of a feature extractor followed by a feature classifier. Specifically, in the conventional deep learning model for person Re-ID, discriminative features $f_a, f_b,f_c \dots$ extracted by the feature extractor, which correspond to the person images of the same person label, would be classified as the same label by the feature classifier. In this way, after it updates the network weight parameter during the training phase, $f_a, f_b,f_c \dots$ would be closer in terms of Euclidean distance or cosine distance. Thus, we can re-identify a person by the Euclidean distance or cosine distance between discriminative features during the test phase. 

In our MLFE, discriminative features $f_{x_i}^{A}$ and $f_{x_i}^{B}$ are both classified by the same classifier in branch $c_i$, and they are classified as the same label $y_i$. According to the above analysis, it actually makes $f_{x_i}^{A}$ and $f_{x_i}^{B}$ closer during the training phase. In other words, with feature $f_{x_i}^{A}$ closing to $f_{x_i}^{B}$ and $f_{x_i}^{B}$ closing to $f_{x_i}^{A}$, it makes the learning information deliver between network $A$ and network $B$ mutually, which indicates that they learn from each other mutually. Our mutual learning is guided by discriminative features; thus it is a ``feature-wise'' method. In the following, we introduce Mutual Learning on Feature Classifiers, where classifier branches learn from each other guided by discriminative features.  
%

\subsection{Mutual Learning on Feature Classifiers (MLFC)}
\label{sec:FCML}

Recalling Per-Camera Learning in the above section, we note that updating the parameter weights of the $C$ classifier branches depends on the different discriminative features corresponding to the $C$ camera views respectively during the training phase, and different classifier branches do not depend on the same discriminative feature in this design. In other words, we adapt feature $f_{x_i}$ to classifier branch ${c_i}$, but we do not adapt $f_{x_i}$ to other classifier branches $k$($k \ne c_i$), which implies that the discrimination ability of different classifier branches would be extremely partial to their corresponding camera view after the training phase. However, we cannot adapt feature $f_{x_i}$ to classifier branch $k$($k \ne c_i$) directly, because the mapping relation of the person label between camera view $c_i$ and camera view $k$($k \ne c_i$) is unknown and the person with label $y_i$ in camera view $c_i$ may not exist in camera view $k$.

To this end, we adapt feature $f_{x_i}$ to other classifier branches in an elegant way during the training phase, which achieves our MLFC. When we leverage $f_{x_i}$ as a discriminative feature input to classifier branch $k$($k \ne c_i$), it outputs a result of predicts of the person in camera view $k$. As analyzed above, we have no knowledge of which person label in camera view $k$ is the correct predict; thus, we cannot propagate the loss during back-propagation within the training phase. Nevertheless, we can feed another feature $f_{x_j}$, where $c_j = c_i$ and $y_i = y_j$, into classifier branch $k$, and obtain another predicts. After that, we pull these two predicts closer, which achieves feature adaption across camera views.

Formally, we define a pair set as follows:  
\begin{equation}
\label{equation:rho}
	\rho= \{(x_i\in \chi ,x_j\in \chi)\mid c_i=c_j, y_i=y_j, i \ne j\}
\end{equation}

We extend the predicts $P^{\Theta}(x_i)$ to $P^{\Theta}(x_i;k)$, which means the predicts of the classifier branch $k$. We leverage Kullback Leibler (KL) Divergence to measure the distance between two predicts of a pair of images $x_i$ and $x_j$ within the branch $k$:



\begin{equation}
\begin{split}
	D_{KL}^{\Theta}(x_i,x_j;k) = \sum_{l=1}^{M_k} P_l^{\Theta}(x_j;k) \times log\frac{P_l^{\Theta}(x_j;k)}{P_l^{\Theta}(x_i;k)} 
\end{split}
\end{equation}
where $P_l^{\Theta}(x_j;k) \in \mathbb{R}$ is the $l^{th}$ element of $P^{\Theta}(x_j;k)$. 

KL Divergence is an asymmetrical distance metric, while a symmetrical distance metric is more appropriate. We adopt Jensen-Shannon (JS) Divergence, a symmetrical distance metric, which is based on KL Divergence.
\begin{equation}
	D_{JS}^{\Theta}(x_i,x_j;k) = \frac{1}{2}D_{KL}^{\Theta}(x_i,x_j;k) + \frac{1}{2}D_{KL}^{\Theta}(x_j,x_i;k) 
\end{equation}

In a mini-batch, we compute the arithmetic mean of all $D_{JS}^{\Theta}(\cdot)$ in the pair set $\rho$, which formulates our MLFC for feature adaption across camera views. Specifically, it is formulated as follows: 
\begin{equation}
	L^{\Theta}_{JL} = \sum_{(x_i,x_j)\in \rho} \sum_{k=1}^{C} \frac{D_{JS}^{\Theta}(x_i,x_j;k)}{C \times \Vert \rho \Vert}
\end{equation}
where $C \times \Vert \rho \Vert$ is a normalization factor.

When it is applied to the two networks design introduced in the last section, we can obtain the following:
\begin{equation}
	L_{JL} = \frac{L^{A}_{JL} + L^{B}_{JL}}{2}
\end{equation}

If MLFC is absent, different classifier branches would not be fed by the same features. With MLFC, different classifier branches would depend on the same features when updating the weight parameters in the training phase, which indicates that learning information would deliver among these classifiers. Namely, they learn from each other mutually by the guidance of these same features, which we also call ``feature-wise'' manner. 

\subsection{Model Training and Testing}
\label{sec:model_train_test}
To summarize, the objective function of our model is formulated as follows:
\begin{equation}
    L = L_{PL} + \lambda L_{CL} + \gamma L_{JL}
\end{equation}
where $\lambda$ and $\gamma$ control the relative importance of the three objective losses, and $L_{CL}$ is expanded from one network to two networks:
\begin{equation}
    L_{CL} = \frac{L_{CL}^{A}+L_{CL}^{B}}{2}
\end{equation}
where the formulations of $L_{CL}^{A}$ and $L_{CL}^{B}$ are similar to that of $L_{CL}^{0}$.

In the testing phase, the final similarity between two image is computed as follows:
\begin{equation}
    s(x_i,x_j) = \frac{cos\_sim(f_{x_i}^{A},f_{x_j}^{A})+cos\_sim(f_{x_i}^{B},f_{x_j}^{B})}{2}
\end{equation}

\section{Experiments}
\label{sec:experiments}

\subsection{Experimental Setting}
\label{sec:experiment_setting}

\subsubsection{Datasets}
\label{sec:dataset}
We evaluate our model on a widely used video-based person Re-ID benchmark dataset MARS~\cite{MARS}, and on two image-based person Re-ID datasets, Market-1501~\cite{market1501} and DukeMTMC-ReID~\cite{Duke}. In our experiment, we adopt the standard train/test splits of these datasets shown in Table~\ref{table:dataset_info}.

\subsubsection{Training and Evaluation Protocol}
\label{sec:protocol}
We assume all person images per ID per camera were drawn from a single pedestrian track, as this assumption is usually used in recent studies~\cite{Gong_unsupervised,TAUDL}. This assumption may not always be true in academic datasets, and we have a discussion in section~\ref{sec:reduced}.

In the training split of MARS, there are 8,298 original tracklets. To obtain WSTs, we combine the tracklets that have the same original person label and the same camera view label in MARS. In the training splits of Market-1501 and DukeMTMC-ReID, there are 12,936 and 16,522 images, respectively, but no original tracklet. We also combine the images that have the same original person label and the same camera view label into one WST. Table~\ref{table:tracklet_info} shows the statistics of the number of WSTs. During the training phase, we do not use any original person labels in these three datasets. 

\begin{table}
\footnotesize
\caption{Basic information statistics of the datasets}
\begin{center}
\begin{tabular}{c|c|c|c|c}
\hline
$Dataset$& \#ID & \#Train & \#Test & \#Image \\ 
\hline
MARS & 1,261 & 625  & 636 & 1,191,003 \\ 
Market-1501 & 1,501 & 751 & 750 & 32,668 \\ 
DukeMTMC-ReID & 1,812 & 702 & 1,110 & 36,411 \\
\hline
\end{tabular}
\end{center}
\label{table:dataset_info}
\end{table}

\begin{table}
\footnotesize
\caption{Statistics of \# Weakly Supervised Tracklet(WST) in the training split}
\begin{center}
	\begin{tabular}{c|c|c|c}
\hline
Dataset&\#images & \#Original Tracklets & \#WST  \\ 
\hline 
MARS &509,914 & 8,298 & 1,955 \\
Market-1501&12,936 & - &  3,262   \\ 
DukeMTMC-ReID&16,522& - & 2,196  \\ 
\hline
\end{tabular}
\end{center}
\label{table:tracklet_info}
\end{table}

In the test splits of MARS, Market-1501 and DukeMTMC-ReID, we adopt the standard single query evaluation protocol in these three datasets, and there are no differences from the common supervised person Re-ID. Mean average precision (mAP) and Rank-1 will be adopted to measure the performance.

\subsection{Implementation Details}
\label{sec:details}

We implemented our model in the PyTorch framework.We adopt ResNet-50~\cite{resnet} pretrained on ImageNet~\cite{imagenet} as network $A$ and network $B$.  All person images are resized to $256 \times 128$ using bilinear interpolation. We set the batch size as B=64 and select these 64 images in a special random way as described below. To ensure that we have an adequate number of images in a single camera view and make $\mu_u$ more accurate in Eq.~(\ref{equation:mu}), we only select images from two camera views in a mini-batch, with 32 images per camera view. To make $\rho$ in Eq.~(\ref{equation:rho}) have more elements, we randomly select 16 pairs of images from per camera view mentioned above separately, for which each pair of images has the same person label. We set weight coefficient $\lambda = 0.2$ and $\gamma = 0.4$. For model optimization, we adopt the ADAM algorithm with an incipient learning rate of $5 \times 10^{-4}$ with the two moment terms $\beta1 = 0.9$ and $\beta2 = 0.999$.

\begin{table}
\footnotesize
\caption{Comparison on Market-1501}
\begin{center}
\begin{tabular}{c|l|c|c}
\hline
Supervision&Method&Rank-1(\%)&mAP(\%) \\
\hline
\multirow{9}{*}{Unsupervised}
&CycleGAN&$48.1$&$20.7$ \\
&PUL&$50.9$&$24.8$ \\
&CAMEL&$54.5$&$26.3$ \\
&TJ-AIDL&$58.2$&$26.5$ \\
&HHL&$62.2$&$31.4$ \\
&TAUDL&$63.7$&$41.2$ \\
&MAR&$67.7$&$40.0$ \\
&UTAL&$69.2$&$46.2$ \\
&ECN&$75.1$&$43.0$ \\
\hline
\multirow{1}{*}{\textbf{Weakly Supervised}}
&\textbf{DFML(ours)}&$81.29$&$56.47$ \\
\hline
\multirow{7}{*}{Supervised} 
&MSCAN&$80.31$&$57.53$ \\
&DF&$81.0$&$63.4$ \\
&SSM&$82.21$&$68.80$ \\
&SVDNet&$82.3$&$62.1$ \\
&GAN&$83.97$&$66.07$ \\
&PCB&$92.4$&$77.3$ \\
\hline
\end{tabular}
\end{center}
\label{table:sota_market}
\end{table}

\subsection{Comparisons with the State-of-the-Art Methods}
\label{sec:compare_SOAT}

We evaluate our model against different kinds of methods, including unsupervised methods(CycleGAN~\cite{cyclegan}, PUL~\cite{PUL}, CAMEL~\cite{CAMEL}, TJ-AIDL~\cite{TJ_AIDL}, HHL~\cite{HHL}, TAUDL~\cite{TAUDL}, MAR~\cite{MAR}, DAL~\cite{DAL}, UTAL~\cite{UTAL}, ECN~\cite{ECN}), weakly supervised methods(UTAL~\cite{UTAL}) and supervised methods(MSCAN~\cite{MSCAN}, DF~\cite{DF}, OIM~\cite{OIM}, QAN~\cite{QAN}, SSM~\cite{SSM}, SVDNet~\cite{SVDNet}, Snippet~\cite{Snippet},PAN~\cite{PAN}, GAN~\cite{GAN}, PCB~\cite{PCB}). 

\begin{table}
\footnotesize
\caption{Comparison on DukeMTMC-ReID}
\begin{center}
\begin{tabular}{c|l|c|c}
\hline
Supervision&Method&Rank-1(\%)&mAP(\%) \\
\hline
\multirow{8}{*}{Unsupervised} 
&PUL&$36.5$&$21.5$ \\
&CycleGAN&$38.5$&$19.9$ \\
&TJ-AIDL&$44.3$&$23.0$ \\
&HHL&$46.9$&$27.2$ \\
&TAUDL&$61.7$&$43.5$ \\
&UTAL&$62.3$&$44.6$ \\
&ECN&$63.3$&$40.4$ \\
&MAR&$67.1$&$48.0$ \\
\hline
\multirow{1}{*}{\textbf{Weakly Supervised}}
&\textbf{DFML(ours)}&$68.17$&$47.12$ \\
\hline
\multirow{5}{*}{Supervised} 
&GAN&$67.68$&$47.13$ \\
&OIM&$68.1$&- \\
&PAN&$71.6$&$51.5$ \\
&SVDNet&$76.7$&$56.8$ \\
&PCB&$81.9$&$65.3$ \\
\hline
\end{tabular}
\end{center}
\label{table:sota_duke}
\end{table}

\begin{table}
\footnotesize
\caption{Comparison on MARS}
\begin{center}
\begin{tabular}{c|l|c|c}
\hline
Supervision&Method&Rank-1(\%)&mAP(\%) \\
\hline
\multirow{3}{*}{Unsupervised} 
&TAUDL&$43.8$&$29.1$ \\
&DAL&$46.8$&$21.4$ \\
&UTAL&$49.9$&$35.2$ \\
\hline
\multirow{2}{*}{\textbf{Weakly Supervised}}
&UTAL(weakly)&$59.5$&$51.7$ \\
&\textbf{DFML(ours)}&$63.91$&$52.02$ \\
\hline
\multirow{2}{*}{Supervised} 
&QAN&$73.7$&$51.7$ \\
&Snippet&$86.3$&$76.1$ \\
\hline
\end{tabular}
\end{center}
\label{table:sota_mars}
\end{table}

We compare our model with 10 state-of-the-art unsupervised methods on three datasets. From results shown in Table~\ref{table:sota_market}, \ref{table:sota_duke}, \ref{table:sota_mars}, our model achieves higher accuracy on both image-based datasets and video-based datasets than those unsupervised methods. On Market-1501 dataset, the best unsupervised method, ECN, is lower than ours by up to 6\% in Rank-1, and the number reaches 14\% for UTAL on MARS dataset. Moreover, most of these unsupervised methods require other annotation data for auxiliary learning, while our weakly supervised annotation only needs little cost.

Compared with supervised learning methods, our method seems still competitive. For example, NSCAN and DF are inferior to ours on Market-1501 dataset, while GAN and OIM are inferior to our method on DukeMTMC-ReID dataset.

Because of the variety of weakly supervised settings, we only compare our method with the latest proposed UTAL whose setting is close to ours. Table~\ref{table:sota_mars} shows that our method is superior to UTAL by 4.4\% in Rank-1 on MARS dataset. The result validates the effectiveness of our method again.

\subsection{Ablation Study}
\label{sec:ablation_study}

\begin{table}
\footnotesize
\caption{Ablation study of MLFE and MLFC on Market-1501}
\begin{center}
\begin{tabular}{c|c|c|c|c}
\hline
method&MLFE&MLFC&Rank-1(\%)&mAP(\%) \\
\hline
\hline
Our Baseline(without $L_{cl}$) &\ding{55}&\ding{55}& $66.33$ & $40.78$ \\
Our Baseline &\ding{55}&\ding{55}& $72.06$ & $44.69$ \\
\hline
\multirow{3}{*}{DFML} 
&\ding{55}&$\checkmark$& $78.74$ & $52.68$ \\
&$\checkmark$&\ding{55}& $72.35$ & $45.76$ \\ 
&$\checkmark$&$\checkmark$& $81.29$ & $56.47$ \\ 
\hline
\end{tabular}
\end{center}
\label{table:ablation_study}
\end{table}



We conduct an ablation experiment to demonstrate the effectiveness of both MLFE and MLFC. 

As shown in Table~\ref{table:ablation_study}, when MLFE is absent, the performance decreases by 2.55\%(81.29\%-78.74\%) in Rank-1 on Market-1501. The main reason for this result is that without two networks restricting and learning from each other, the feature learned by one network would adapt to their corresponding camera view space in haste. Therefore the general view information will be weaker, which certainly decreases the performance. 

On the other hand, when MLFC is absent, the performance decreases by 8.94\%(81.29\%-72.35\%). The main reason is that without feature adaption across different camera views, the Re-ID spaces of different camera views do not fuse very well.





\subsection{Further Analysis}
\label{sec:further_analysis}
%
\subsubsection {Reduced Dataset}
\label{sec:reduced}

In most circumstances of the real-world, a persons pass by a camera view only once withing a very long time(or never return), and it is approximately true on DukeMTMC-ReID but there are some exceptions in Market-1501 and MARS. Market-1501 and MARS are originated from the same source videos, thus here we only have a discussion on Market-1501. 

To make Market-1501 more realistic, we reduce Market- 1501 in the following way: if the time difference of two person images that belong to the same camera view and the same person is within x minutes(e.g. 3 minutes or 5 minutes), we assume they belong to the same WST. Then we have one or more WSTs per person per camera view. We reserve the longest WST per person per camera view, and remove the other WST. In this way, we have two datasets named Market-1501-3min-WST and Market-1501-5min-WST respectively, which is shown in Table~\ref{table:xmintable_info}.

Our experiment on these two datasets, which is shown in Table~\ref{table:xmintable_performance}, indicated that we still got a considerable performance of Rank-1 74.91\% and Rank-1 77.08\%. Note that the number of person images of these two datasets is just 77.8\% and 81.64\% of the original Market-1501. This result validates the effectiveness of our method again.

\subsubsection {More Networks on MLFE}
\label{sec:more_FEML}
We explore the use of more than two networks in MLFE. As shown by the experiment results in Table~\ref{table:more_FEML}, compared with the two networks design in MLFE, the three networks design achieves the similar performance, which indicates that too many networks restricting each other in MLFE may not lead to better performance. Therefore, two networks design is sufficient in our MLFE.

\begin{table}
\footnotesize
\caption{Comparison with more networks on Market-1501}
\begin{center}
\begin{tabular}{c|c|c}
\hline
method&Rank-1(\%)&mAP(\%) \\
\hline
\hline
	DFML(without MLFE,one network) & 78.74&52.68 \\
DFML(two networks) & 81.29 & 56.47 \\ 
DFML(three networks) &81.08 & 57.86 \\
\hline
\end{tabular}
\end{center}
\label{table:more_FEML}
\end{table}

\begin{table}
\footnotesize
\caption{\#images of Market1501-3min-WST and Market1501-5min-WST in the training split}
\begin{center}
\begin{tabular}{c|c|c}
\hline
	DataSet&\#images&remaining rate(\%)  \\
\hline
	Market-1501&12,936&$100$ \\
	Market-1501-3min-WST&10,065&$77.8$ \\
	Market-1501-5min-WST&10,564&$81.64$ \\
\hline
\end{tabular}
\end{center}
\label{table:xmintable_info}
\end{table}

\begin{table}
\footnotesize
\caption{performance of Market1501-3min-WST and Market1501-5min-WST}
\begin{center}
\begin{tabular}{c|c|c}
\hline
	DataSet&Rank-1(\%)&mAP(\%)  \\
\hline
	Market-1501-3min-WST&74.91&$49.08$ \\
	Market-1501-5min-WST&77.08&$52.26$ \\
\hline
\end{tabular}
\end{center}
\label{table:xmintable_performance}
\end{table}




\subsubsection{Large Number of Camera Views}
\label{sec:large_number_camera}
When there are too many camera views in the training dataset, MLFC would clearly be slow because we should adapt the discriminative feature to too many classifiers of other camera views. To mitigate this problem, for a person image of camera view $i$, we only adapt the respective feature to the classifier of the camera view $k$ in which the person occurred in camera view $i$ usually occurs. For example, camera $A$ is quite close to camera $B$ but far away from camera $C$ in the real world. Persons occurred in camera view A would usually occur in camera view B but not camera view C; thus, we could adapt the discriminative feature from camera $A$ to the classifier of camera $B$ but not camera $C$. Specifically, we define the following probability function:
\begin{equation}
	\phi(i,j) = P(occur\ in\ view\ i | occur\ in\ view\ j)
\end{equation}
Briefly, $\phi(i,j)$ is the probability that a person occurs in camera view $i$ in the condition of this person occurring in camera view $j$. Then we rewrite MLFC formulated as follows: 

\begin{equation}
	L^{\Theta}_{JLM} = \sum_{(x_i,x_j)\in \rho} \sum_{k=1 }^{C} \frac{\mathbbm{1}(\phi(k,c_i) \ge \eta) D_{JS}^{\Theta}(x_i,x_j;k)}{\Vert \rho \Vert \times \sum_{k=1 }^{C} \mathbbm{1}(\phi(k,c_i) \ge \eta) }
\end{equation}
where $\eta$ is the probability threshold and $\Vert \rho \Vert \times \sum_{k=1 }^{C} \mathbbm{1}(\phi(k,c_i) \ge \eta)$ is a normalization factor. Specially, $\eta =0$ means we adopt the original MLFC, and $\eta=1$ means MLFC is absent. 
\begin{equation}
	L_{JLM} = \frac{L^{A}_{JLM} + L^{B}_{JLM}}{2}
\end{equation}
In this way, we obtain the loss function $L_{JLM}$ of the formulation where we only adapt the discriminative feature of person image to the classifier of the camera view in which this person occurs with a certain probability.

As a matter of fact, it is difficult to obtain the exact value of $\phi(\cdot)$ in the real world. We can roughly estimate it according to some indirect factor, such as the distance between cameras. To demonstrate how our MLFC works, we use the fully supervised annotation of datasets to obtain the exact value of $\phi(\cdot)$ in the following experiment.

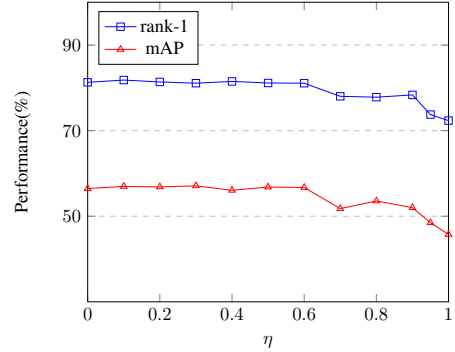
\begin{figure}
\centering
\begin{tikzpicture}[scale=0.7]
\begin{axis}[
    xlabel={$\eta$},
	ylabel={Performance(\%)},
    xmin=0, xmax=1,
    ymin=30, ymax=100,
    xtick={0,0.2,0.4,0.6,0.8,1.0},
    ytick={50,70,90},
    legend pos=north west,
    ymajorgrids=true,
    grid style=dashed,
]
 
\addplot[
    color=blue,
    mark=square,
    ]
    coordinates {
	    (0,81.29) (0.1,81.82) (0.2, 81.38) (0.3, 81.09) (0.4,81.50) (0.5,81.14) (0.6,81.08) (0.7,78.02) (0.8,77.82) (0.9, 78.35) (0.95, 73.75) (1,72.35)
    };
    \addlegendentry{rank-1}
\addplot[
    color=red,
    mark=triangle,
    ]
    coordinates {
	    (0,56.47) (0.1,56.97) (0.2,56.86) (0.3, 57.10) (0.4,56.06) (0.5,56.82) (0.6,56.72) (0.7,51.78) (0.8,53.56) (0.9, 52.01) (0.95, 48.49) (1,45.76)
    };
    \addlegendentry{mAP}
 
\end{axis}
\end{tikzpicture}
\caption{Performance of various $\eta$ on Market-1501}
\label{fig:moreview}
\end{figure}

In contrast to our full model, we replace $L_{JL}$ with $L_{JLM}$ in this experiment. The experiment shown in Fig.~\ref{fig:moreview} reveals the performance sensitivity of threshold $\eta$ on Market-1501, from which we observe that we can also obtain a high performance in our model where $\eta$ is approximately less than 0.6. Thus, in real-world applications, it is unnecessary for a discriminative feature to adapt to every classifier branch, and we only adapt the discriminative feature of a person image to the classifier of the camera views in which this person occurs usually.

\section{Conclusion}
In this work, we proposed a novel Deep Feature-wise Mutual Learning (DFML) model for weakly supervised tracklet person re-identification. This model learns from Weakly Supervised Tracklet (WST) data, obtained from combining the fragmented video person tracklets of the same person in the same camera view over a period of time. In contrast to fully supervised Re-ID data, our WST is quite cheaper to obtain, which enables our DFML to be more scalable in real-world applications. Our proposed DFML consists of Mutual Learning on Feature Extractors (MLFE) and Mutual Learning on Feature Classifiers (MLFC), where MLFC makes feature more adaptive to each camera view and MLFE makes feature more robust and discriminative. Extensive experiments demonstrate the superiority of our proposed DFML over the state-of-the-art unsupervised models, and even some supervised Re-ID models.

{\small
\bibliographystyle{ieee_fullname}
\bibliography{mypaper}

\begin{thebibliography}{10}\itemsep=-1pt

\bibitem{ahmed2015improved}
Ejaz Ahmed, Michael Jones, and Tim~K Marks.
\newblock An improved deep learning architecture for person re-identification.
\newblock In {\em Proceedings of the IEEE conference on computer vision and
  pattern recognition}, pages 3908--3916, 2015.

\bibitem{SSM}
Song Bai, Xiang Bai, and Qi Tian.
\newblock Scalable person re-identification on supervised smoothed manifold.
\newblock In {\em Proceedings of the IEEE Conference on Computer Vision and
  Pattern Recognition}, pages 2530--2539, 2017.

\bibitem{Snippet}
Dapeng Chen, Hongsheng Li, Tong Xiao, Shuai Yi, and Xiaogang Wang.
\newblock Video person re-identification with competitive snippet-similarity
  aggregation and co-attentive snippet embedding.
\newblock In {\em Proceedings of the IEEE Conference on Computer Vision and
  Pattern Recognition}, pages 1169--1178, 2018.

\bibitem{DAL}
Yanbei Chen, Xiatian Zhu, and Shaogang Gong.
\newblock Deep association learning for unsupervised video person
  re-identification.
\newblock {\em arXiv preprint arXiv:1808.07301}, 2018.

\bibitem{imagenet}
Jia Deng, Wei Dong, Richard Socher, Li-Jia Li, Kai Li, and Li Fei-Fei.
\newblock Imagenet: A large-scale hierarchical image database.
\newblock In {\em Proceedings of the IEEE Conference on Computer Vision and
  Pattern Recognition}, pages 248--255, 2009.

\bibitem{SPGAN}
Weijian Deng, Liang Zheng, Qixiang Ye, Guoliang Kang, Yi Yang, and Jianbin
  Jiao.
\newblock Image-image domain adaptation with preserved self-similarity and
  domain-dissimilarity for person re-identification.
\newblock In {\em Proceedings of the IEEE Conference on Computer Vision and
  Pattern Recognition}, pages 994--1003, 2018.

\bibitem{PUL}
Hehe Fan, Liang Zheng, Chenggang Yan, and Yi Yang.
\newblock Unsupervised person re-identification: Clustering and fine-tuning.
\newblock {\em ACM Transactions on Multimedia Computing, Communications, and
  Applications (TOMM)}, 14(4):83, 2018.

\bibitem{resnet}
Kaiming He, Xiangyu Zhang, Shaoqing Ren, and Jian Sun.
\newblock Deep residual learning for image recognition.
\newblock In {\em Proceedings of the IEEE Conference on Computer Vision and
  Pattern Recognition}, pages 770--778, 2016.

\bibitem{distillation}
Geoffrey Hinton, Oriol Vinyals, and Jeff Dean.
\newblock Distilling the knowledge in a neural network.
\newblock {\em arXiv preprint arXiv:1503.02531}, 2015.

\bibitem{BN}
Sergey Ioffe and Christian Szegedy.
\newblock Batch normalization: Accelerating deep network training by reducing
  internal covariate shift.
\newblock {\em arXiv preprint arXiv:1502.03167}, 2015.

\bibitem{MSCAN}
Dangwei Li, Xiaotang Chen, Zhang Zhang, and Kaiqi Huang.
\newblock Learning deep context-aware features over body and latent parts for
  person re-identification.
\newblock In {\em Proceedings of the IEEE Conference on Computer Vision and
  Pattern Recognition}, pages 384--393, 2017.

\bibitem{TAUDL}
Minxian Li, Xiatian Zhu, and Shaogang Gong.
\newblock Unsupervised person re-identification by deep learning tracklet
  association.
\newblock In {\em Proceedings of the European Conference on Computer Vision
  (ECCV)}, pages 737--753, 2018.

\bibitem{Gong_unsupervised}
Minxian Li, Xiatian Zhu, and Shaogang Gong.
\newblock Unsupervised tracklet person re-identification.
\newblock {\em IEEE transactions on pattern analysis and machine intelligence},
  2019.

\bibitem{siamese1}
Wei Li, Rui Zhao, Tong Xiao, and Xiaogang Wang.
\newblock Deepreid: Deep filter pairing neural network for person
  re-identification.
\newblock In {\em Proceedings of the IEEE conference on computer vision and
  pattern recognition}, pages 152--159, 2014.

\bibitem{QAN}
Yu Liu, Junjie Yan, and Wanli Ouyang.
\newblock Quality aware network for set to set recognition.
\newblock In {\em Proceedings of the IEEE Conference on Computer Vision and
  Pattern Recognition}, pages 5790--5799, 2017.

\bibitem{weak_supervised1}
Jingke Meng, Sheng Wu, and Wei-Shi Zheng.
\newblock Weakly supervised person re-identification.
\newblock In {\em Proceedings of the IEEE Conference on Computer Vision and
  Pattern Recognition}, pages 760--769, 2019.

\bibitem{Duke}
Ergys Ristani, Francesco Solera, Roger Zou, Rita Cucchiara, and Carlo Tomasi.
\newblock Performance measures and a data set for multi-target, multi-camera
  tracking.
\newblock In {\em Proceedings of the European Conference on Computer Vision
  (ECCV)}, pages 17--35, 2016.

\bibitem{SVDNet}
Yifan Sun, Liang Zheng, Weijian Deng, and Shengjin Wang.
\newblock Svdnet for pedestrian retrieval.
\newblock In {\em Proceedings of the IEEE International Conference on Computer
  Vision}, pages 3800--3808, 2017.

\bibitem{PCB}
Yifan Sun, Liang Zheng, Yi Yang, Qi Tian, and Shengjin Wang.
\newblock Beyond part models: Person retrieval with refined part pooling (and a
  strong convolutional baseline).
\newblock In {\em Proceedings of the European Conference on Computer Vision
  (ECCV)}, pages 480--496, 2018.

\bibitem{UTAL}
Jiajie Tian, Zhu Teng, Rui Li, Yan Li, Baopeng Zhang, and Jianping Fan.
\newblock Imitating targets from all sides: An unsupervised transfer learning
  method for person re-identification.
\newblock {\em arXiv preprint arXiv:1904.05020}, 2019.

\bibitem{weak_supervised2}
Guangrun Wang, Guangcong Wang, Xujie Zhang, Jianhuang Lai, and Liang Lin.
\newblock Weakly supervised person re-identification: Cost-effective learning
  with {A} new benchmark.
\newblock {\em arXiv preprint arXiv:1904.03845}, 2019.

\bibitem{TJ_AIDL}
Jingya Wang, Xiatian Zhu, Shaogang Gong, and Wei Li.
\newblock Transferable joint attribute-identity deep learning for unsupervised
  person re-identification.
\newblock In {\em Proceedings of the IEEE Conference on Computer Vision and
  Pattern Recognition}, pages 2275--2284, 2018.

\bibitem{PTGAN}
Longhui Wei, Shiliang Zhang, Wen Gao, and Qi Tian.
\newblock Person transfer gan to bridge domain gap for person
  re-identification.
\newblock In {\em Proceedings of the IEEE Conference on Computer Vision and
  Pattern Recognition}, pages 79--88, 2018.

\bibitem{OIM}
Tong Xiao, Shuang Li, Bochao Wang, Liang Lin, and Xiaogang Wang.
\newblock Joint detection and identification feature learning for person
  search.
\newblock In {\em Proceedings of the IEEE Conference on Computer Vision and
  Pattern Recognition}, pages 3415--3424, 2017.

\bibitem{siamese2}
Dong Yi, Zhen Lei, Shengcai Liao, and Stan~Z Li.
\newblock Deep metric learning for person re-identification.
\newblock In {\em 2014 22nd International Conference on Pattern Recognition},
  pages 34--39, 2014.

\bibitem{CAMEL}
Hong-Xing Yu, Ancong Wu, and Wei-Shi Zheng.
\newblock Cross-view asymmetric metric learning for unsupervised person
  re-identification.
\newblock In {\em Proceedings of the IEEE International Conference on Computer
  Vision}, pages 994--1002, 2017.

\bibitem{MAR}
Hong-Xing Yu, Wei-Shi Zheng, Ancong Wu, Xiaowei Guo, Shaogang Gong, and
  Jian-Huang Lai.
\newblock Unsupervised person re-identification by soft multilabel learning.
\newblock In {\em Proceedings of the IEEE Conference on Computer Vision and
  Pattern Recognition}, pages 2148--2157, 2019.

\bibitem{mutual}
Ying Zhang, Tao Xiang, Timothy~M Hospedales, and Huchuan Lu.
\newblock Deep mutual learning.
\newblock In {\em Proceedings of the IEEE Conference on Computer Vision and
  Pattern Recognition}, pages 4320--4328, 2018.

\bibitem{DF}
Liming Zhao, Xi Li, Yueting Zhuang, and Jingdong Wang.
\newblock Deeply-learned part-aligned representations for person
  re-identification.
\newblock In {\em Proceedings of the IEEE International Conference on Computer
  Vision}, pages 3219--3228, 2017.

\bibitem{MARS}
Liang Zheng, Zhi Bie, Yifan Sun, Jingdong Wang, Chi Su, Shengjin Wang, and Qi
  Tian.
\newblock Mars: A video benchmark for large-scale person re-identification.
\newblock In {\em Proceedings of the European Conference on Computer Vision
  (ECCV)}, pages 868--884, 2016.

\bibitem{market1501}
Liang Zheng, Liyue Shen, Lu Tian, Shengjin Wang, Jingdong Wang, and Qi Tian.
\newblock Scalable person re-identification: A benchmark.
\newblock In {\em Proceedings of the IEEE international conference on computer
  vision}, pages 1116--1124, 2015.

\bibitem{PRW}
Liang Zheng, Hengheng Zhang, Shaoyan Sun, Manmohan Chandraker, Yi Yang, and Qi
  Tian.
\newblock Person re-identification in the wild.
\newblock In {\em Proceedings of the IEEE Conference on Computer Vision and
  Pattern Recognition}, pages 1367--1376, 2017.

\bibitem{GAN}
Zhedong Zheng, Liang Zheng, and Yi Yang.
\newblock Unlabeled samples generated by gan improve the person
  re-identification baseline in vitro.
\newblock In {\em Proceedings of the IEEE International Conference on Computer
  Vision}, pages 3754--3762, 2017.

\bibitem{PAN}
Zhedong Zheng, Liang Zheng, and Yi Yang.
\newblock Pedestrian alignment network for large-scale person
  re-identification.
\newblock {\em IEEE Transactions on Circuits and Systems for Video Technology},
  2018.

\bibitem{HHL}
Zhun Zhong, Liang Zheng, Shaozi Li, and Yi Yang.
\newblock Generalizing a person retrieval model hetero-and homogeneously.
\newblock In {\em Proceedings of the European Conference on Computer Vision
  (ECCV)}, pages 172--188, 2018.

\bibitem{ECN}
Zhun Zhong, Liang Zheng, Zhiming Luo, Shaozi Li, and Yi Yang.
\newblock Invariance matters: Exemplar memory for domain adaptive person
  re-identification.
\newblock In {\em Proceedings of the IEEE Conference on Computer Vision and
  Pattern Recognition}, pages 598--607, 2019.

\bibitem{cyclegan}
Jun-Yan Zhu, Taesung Park, Phillip Isola, and Alexei~A Efros.
\newblock Unpaired image-to-image translation using cycle-consistent
  adversarial networks.
\newblock In {\em Proceedings of the IEEE international conference on computer
  vision}, pages 2223--2232, 2017.

\end{thebibliography}
}
\end{document}